\documentclass[letterpaper,twocolumn,10pt]{article}
\usepackage{usenix-2020-09}
\usepackage{graphicx}
\usepackage{listings}
\usepackage{booktabs}
\usepackage[linesnumbered,ruled,noend,figure]{algorithm2e}
\usepackage{caption}
\usepackage{subcaption}
\usepackage{amsmath}
\usepackage{fancyhdr}
\usepackage{verbatim}
\usepackage{enumitem}
\usepackage{authblk}

 \usepackage{etoolbox}
 \makeatletter
 \patchcmd{\maketitle}
 	{\@maketitle}
 	{\vspace{-5em}\@maketitle\vspace{-3em}}
 	{}
 	{}
 \makeatother

\newcommand{\oursys}{{A}ttention{E}ngine}

\newcommand{\para}[1]{{\vspace{2pt} \bf \noindent #1 \hspace{1pt}}}

\begin{document}

\hyphenpenalty=600

\lstset{
  language=Python,
  basicstyle=\fontsize{7.5}{8.5}\ttfamily,
  keywordstyle=\color{blue},
  commentstyle=\itshape\color{green!40!black},
  frame=single,
  numbers=left,
  stepnumber=1,
  emph={getGridDim, getBlockDim, kernel, getBlockIdx, StreamModule, virtual, Schedule, Signal, GetBlockExecutorId},
  emphstyle=\textbf,
  morekeywords={class, void, size_t, vector, Map, sEUType},
  deletekeywords={bool},
  emph={[2] StreamModule, virtual, Schedule, Signal, GetBlockExecutorId},
  emphstyle={[2]\color{purple!80!black}}
}

\date{}

\title{\oursys{}: A Versatile Framework for Efficient Attention Mechanisms on Diverse Hardware Platforms}
\renewcommand\footnotemark{}
\author[1,3]{Feiyang Chen}
\author[2,3]{Yu Cheng}
\author[2,3]{Lei Wang}
\author[3]{Yuqing Xia}
\author[3]{Ziming Miao}
\author[3]{Lingxiao Ma}
\author[3]{Fan Yang}
\author[3]{Jilong Xue}
\author[2]{Zhi Yang}
\author[3]{Mao Yang}
\author[1]{Haibo Chen}

\affil[ ]{%
    \textsuperscript{1}Shanghai Jiao Tong University, 
    \textsuperscript{2}Peking University, 
    \textsuperscript{3}Microsoft Research
}

\maketitle

\begin{abstract}

Transformers and large language models (LLMs) have revolutionized machine learning, with attention mechanisms at the core of their success. As the landscape of attention variants expands, so too do the challenges of optimizing their performance, particularly across different hardware platforms. Current optimization strategies are often narrowly focused, requiring extensive manual intervention to accommodate changes in model configurations or hardware environments.

In this paper, we introduce \oursys{}, a comprehensive framework designed to streamline the optimization of attention mechanisms across heterogeneous hardware backends. By decomposing attention computation into modular operations with customizable components, \oursys{} enables flexible adaptation to diverse algorithmic requirements. The framework further automates kernel optimization through a combination of programmable templates and a robust cross-platform scheduling strategy. Empirical results reveal performance gains of up to 10× on configurations beyond the reach of existing methods. \oursys{} offers a scalable, efficient foundation for developing and deploying attention mechanisms with minimal manual tuning. Our code has been open-sourced and is available at \href{https://github.com/microsoft/AttentionEngine}{https://github.com/microsoft/AttentionEngine}.
\end{abstract}

\section{Introduction}
\label{sec:intro}


Attention is a fundamental mechanism in modern large langauge models (LLMs), enabling groundbreaking advancements in natural language understanding and related domains. By dynamically weighting interactions across input tokens, attention allows models to capture sophisticated contextual relationships, making it an indispensable component of modern deep learning systems.

Attention mechanisms dominate the computational workload in LLMs, and their 
proportion continuously increases with the growing sequence length. This trend underscores the critical importance of optimizing attention for end-to-end model training and inference. 
For instance, as illustrated in Table~\ref{tab:llama-proportion}, attention accounts for \(55\%\) of the computational time in LLAMA-3B when the sequence length is 2048. This proportion further escalates to \(82\%\) as the sequence length extends to 8192. 
Such a significant computational burden highlights the necessity of efficient attention mechanisms to ensure optimal performance and scalability of LLMs across various applications and hardware platforms.

However, attention optimization is nontrivial due to high computation and memory demands and often relies on handcrafted kernels. For example, FlashAttention~\cite{dao2022flashattention} employs online softmax, memory-efficient pipelining, and kernel fusion to improve canonical attention; while Mamba2~\cite{dao2024mamba2}, a linear version of attention, utilizes Triton-based~\cite{triton} kernels with selective gating and chunk-based processing for performance improvement. These handcrafted optimizations are labor-intensive, hardware-specific, and constrained to fixed configurations, thus limiting the adaptability to diverse attention designs and configurations.


The diversity of attention variants continues to expand, driven by task-specific requirements and innovations. For instance, sigmoid attention\cite{ramapuram2024sigmoidattn} replaces softmax with sigmoid activation for improved efficiency, and linear attention mechanisms, such as Mamba\cite{dao2024mamba2}, reformulate computation with selective gating for enhanced efficiency. Other variants, like DeepSeek V2\cite{deepseekai2024deepseekv2strongeconomicalefficient} and RetNet\cite{sun2023retentive}, deviate further by requiring non-standard tensor dimensions, introducing additional computational challenges.

Adapting to this growing diversity requires significant expert efforts for kernel customization. Furthermore, differences in Attention input configurations and hardware platforms, such as NVIDIA A100, H100, and AMD MI300X GPUs, complicate the landscape. Hardware differences in tile sizes, memory hierarchies, and pipelining strategies necessitate new implementations, significantly increasing development overheads and limiting scalability. 
For example, FlashAttention v2 reached \(70\%\) of the peak computation throughput on the NVIDIA A100, but only achieved \(30\%\) on the NVIDIA H100. Complex techniques such as register-level pipelining and ping-pong kernel design must be used to achieve peak performance\cite{shah2024flashattention}.

\begin{figure*}[t]
    \centering
    \includegraphics[width=0.8\linewidth]{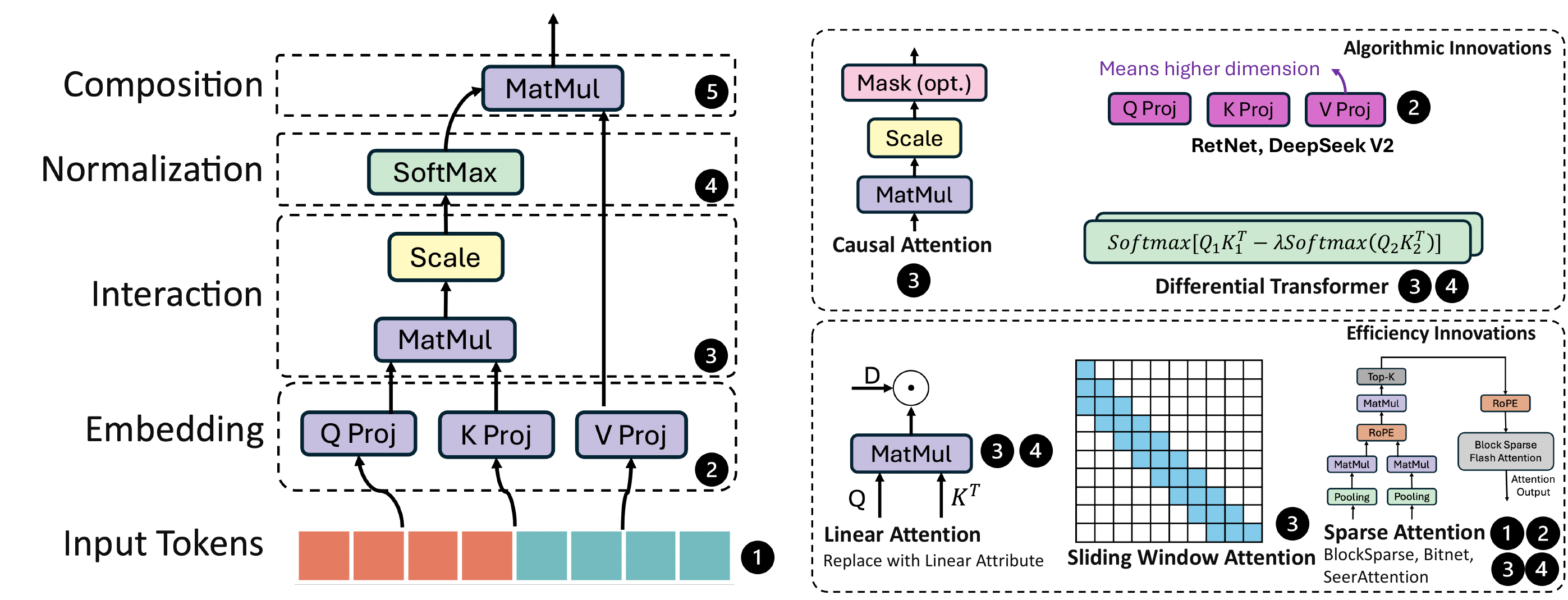}
  \vspace{-2mm}
    \caption{The foundational attention mechanism and its variants: Attention
mechanisms is divided into stages such as embedding, interaction, normalization, and composition(left). Attention variants make various changes to these stages(right). For example, Causal Attention modified the interaction stage to apply a mask, which makes the computation flow different.}
  \vspace{-5mm}
    \label{fig:attention_variant}
\end{figure*}

To address these challenges, we propose \oursys{}, a unified framework for designing, optimizing, and executing diverse attention mechanisms across hardware platforms. At its core, \oursys{} abstracts attention mechanisms into two fundamental operations: relevance scoring and aggregation. These operations capture the essence of attention mechanisms, ensuring a consistent yet flexible foundation for diverse designs.

Building on this abstraction, \oursys{} introduces customizable attention templates that fix the core operations of relevance scoring and aggregation while exposing \texttt{customizable functions} for user-defined extensions. These functions allow users to design their attention variants by applying transformations like masking, scaling, or row-wise normalization, enabling seamless adaptation to task-specific requirements.

One challenge is how to retain high performance customization despite abstraction. \oursys{} enables automated optimization through a cross-backend scheduling and execution framework that dynamically adapts to input configurations and hardware constraints. By abstracting kernel generation and optimization complexities, \oursys{} supports a wide range of attention variants and hardware platforms while delivering exceptional performance.

We implemented \oursys{} with 7.3k lines of C++ and Python code and have open-sourced the system to foster further innovations. Evaluation results demonstrate that \oursys{} achieves performance comparable to handcrafted expert-optimized kernels, delivering up to 10.4× speedup for configurations unsupported by existing implementations. Moreover, \oursys{} provides unparalleled flexibility for designing and optimizing custom attention mechanisms, marking a significant step toward scalable and generalizable attention computation.


\section{Background}
\label{sec:moti}






\subsection{Attention Mechanisms}

Large Language Models (LLMs) have transformed natural language processing (NLP), enabling breakthroughs in tasks such as text understanding and generation. At the core of this success is the attention mechanism, which allows models to selectively focus on relevant parts of an input sequence, significantly enhancing sequence-to-sequence tasks like translation~\cite{bahdanau2014neural}. Attention computes pairwise relevance, or \textit{attention scores}, between input tokens, which are then used to weight and aggregate token representations, guiding the generation of output tokens.

The introduction of Queries (Q), Keys (K), and Values (V) in the Transformer architecture~\cite{Ashish17AttentionIsAllYouNeed} formalized and generalized attention computation. Queries represent what the model seeks, Keys encode the input attributes, and Values carry the associated content. Modern attention computation, as summarized in Figure~\ref{fig:attention_variant}, follows five key stages:
\begin{itemize}[noitemsep, topsep=0pt]
    \item \textbf{Input Tokens:} Raw input sequences serve as the foundation for computation.
    \item \textbf{Embedding:} Input tokens are mapped to continuous vector representations through projections of Q, K, and V matrices, encapsulating semantic information.
    \item \textbf{Interaction:} Pairwise relevance scores are computed using the dot product of Q and K, optionally scaled, to quantify token relationships.
    \item \textbf{Normalization:} Relevance scores are transformed into normalized weights using functions like softmax, ensuring interpretability and row-wise consistency.
    \item \textbf{Composition:} Weighted scores are combined with V representations to generate context vectors, integrating information from relevant tokens into a single output for each token.
\end{itemize}

The Q, K, V framework has established itself as the foundation of modern attention mechanisms, offering scalability, flexibility, and computational efficiency. This structured approach underpins the success of neural architectures in addressing a wide range of NLP tasks.

\subsection{Diversity in Attention Mechanisms}
 
Building on the foundational design of attention mechanisms, researchers have introduced numerous variants aimed at improving performance, addressing task-specific requirements, and enhancing computational efficiency. As illustrated on the right side of Figure~\ref{fig:attention_variant}, these innovations can be categorized into algorithmic and efficiency advancements, each targeting specific stages of the attention mechanism. 

\para{Algorithmic Innovations}
  focus on enhancing robustness, accuracy, and task-specific capabilities in attention:
\begin{itemize}[noitemsep,topsep=0pt, left=0pt]
    \item Task-Specific Modifications: Causal attention\cite{Ashish17AttentionIsAllYouNeed} modifies the interaction stage by restricting interactions to prior tokens. This design supports autoregressive decoding, a critical feature for applications like text generation and speech synthesis.
    \item Improved Robustness and Accuracy: DiffTransformer\cite{ye2024differentialtransformer} refines both the interaction and normalization stages for higher accuracy and reduced noise in attention scores.
    \item Non-Conventional Tensor Dimensions: Models like DeepSeek V2\cite{deepseekai2024deepseekv2strongeconomicalefficient} and RetNet\cite{sun2023retentive} enhance the embedding stage by employing higher hidden dimensions, enabling richer semantic representation.
\end{itemize}

\para{Efficiency Innovations}
aim to reduce computational overhead while maintaining the effectiveness of attention:
\begin{itemize}[noitemsep,topsep=0pt, left=0pt]
    \item Compact Representations: Linear attention, such as Mamba\cite{dao2024mamba2} and the recurrent form of RetNet\cite{sun2023retentive}, transforms the interaction, normalization and composition stages by compressing past information into compact $KV$ representations. Sliding Window Attention\cite{beltagy2020longformerlongdocumenttransformer} modifies the interaction stage by limiting the attention scope to a fixed local window, optimizing memory usage and computational focus.
    \item Sparse Attention: Sparse attention mechanisms, such as BigBird\cite{zaheer2021bigbirdtransformerslonger}, SeerAttention\cite{gao2024seerattentionlearningintrinsicsparse}, and BitNet\cite{wang2023bitnetscaling1bittransformers}, introduce sparsity across multiple stages, including input tokens, embedding, interaction, and normalization. These methods leverage structured patterns or treat low-bit precision as sparse regions to reduce computational and memory demands without sacrificing effectiveness.
\end{itemize}

\begin{table}[t]
  \centering
  \resizebox{0.3\textwidth}{!}{
    \begin{tabular}{l|c|c|c|c}
    \hline
    Seqlen & 2K    & 4K    & 6K    & 8K \\
    \hline
    LLAMA-3B &   55\%    &   70\%    &   78\%    & 82\% \\
    \hline
    \end{tabular}%
    }
  \caption{Attention proportion in LLAMA-3B inference}
  \vspace{-5mm}
  \label{tab:llama-proportion}%
\end{table}%

\subsection{Efficient Implementation of Attention}
The attention mechanism takes large proportion in LLM computation. Table~\ref{tab:llama-proportion} shows the attention proportion in LLAMA-3B inference.
Efficient implementations of various attention mechanisms hinge on reducing memory access and maximizing the utilization of compute units. Many libraries with handcrafted kernels achieve this by fusing memory-intensive operations, including element-wise calculations and reductions.

\begin{figure}[t]
    \centering
    \includegraphics[width=0.8\linewidth]{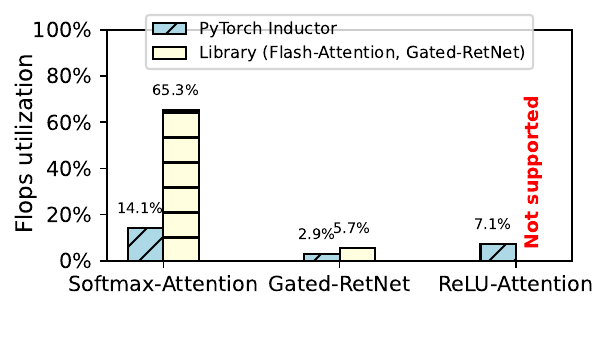}
    \vspace{-0.7cm}
    \caption{The performance of attention implementations.}
  \vspace{-5mm}
    \label{fig:torch-vs-handcraft}
\end{figure}

FlashAttention~\cite{shah2024flashattention} exemplifies this approach by integrating softmax computation, memory-efficient pipelining, and kernel fusion, thereby reducing computational overhead and improving performance. However, these libraries impose strict constraints on the attention patterns they support. Even minor deviations, such as the atypical input dimensions used in DeepSeek V2 and RetNet, can invalidate these optimizations. Figure~\ref{fig:torch-vs-handcraft} illustrates the performance disparity across different attention variants. For standard Softmax-Attention, the handcrafted library FlashAttention3~\cite{shah2024flashattention} significantly outperforms the native PyTorch implementation, achieving over 60\% FLOPS utilization. In contrast, for less common variants like Gated-RetNet and ReLU-Attention, these libraries exhibit poor performance or provide no support at all.

Additionally, due to limited development resources, these libraries predominantly target top-tier hardware, such as NVIDIA's H100 and A100 GPUs, and are not easily transferable to alternative platforms like AMD GPUs. Adapting these implementations to different hardware ecosystems remains challenging and demands significant expertise.

To simplify kernel development, automated compilers\cite{24pytorch2, tvm2018, ansor, xla, shi2023welder, tensorrt} have emerged. While these tools reduce development effort, they struggle to match the performance of handcrafted kernels for attention variants. This limitation arises from their inability to fully understand the semantics of attention computation, as they often treat it as a sequence of discrete and opaque operations. Advanced optimizations, such as transforming softmax into an online softmax, are beyond the scope of current compiler capabilities, resulting in suboptimal performance.

To balance performance and development efficiency, some approaches adopt trade-offs between flexibility and optimization. For instance, FlexAttention~\cite{dong2024flexattentionprogrammingmodel} utilizes a template-based methodology in which the majority of the computation is predefined, while exposing a limited set of customizable functions to users. This design enables the optimization of the entire attention operation while providing some flexibility for specific variants. However, these templates are derived from the computational flow of a particular variant, making it difficult to generalize to a wider range of attention variants, such as linear attention.

\section{A Unified Attention Abstraction}
\label{sec:attn}
Attention mechanisms exhibit significant diversity at the implementation level. For example, standard attention utilizes matrix multiplication to compute attention scores between $Q$ and $K$, followed by a weighted aggregation of $V$ to produce the output representation. In contrast, linear attention compresses $K$ and $V$ using a recurrent loop before applying $Q$ to compute the output. Despite these implementation differences, these variants adhere to the same underlying principles of attention semantics.

By examining the native implementation of attention as a loop-based operation, we identify two fundamental components common to all attention mechanisms:

\begin{itemize}[noitemsep,topsep=0pt, left=0pt]
    \item \textit{Relevance Scoring:} This operation forms the core of attention mechanisms, capturing pairwise similarities or interactions between input tokens. It is typically realized through inner products or other similarity measures to determine token relationships.
    \item \textit{Aggregation:} Using the relevance scores, this operation consolidates contextual information into a representation for each token. 
\end{itemize}

Building on these two fundamental operations, we propose a unified template that encapsulates the diverse spectrum of attention variants. This template abstracts the core semantics of relevance scoring and aggregation while offering customizable components, striking a balance between broad applicability and development flexibility. By providing a consistent framework, this approach streamlines the design and implementation of new attention mechanisms while enabling efficient adaptation to evolving computational demands. The next section introduces \oursys{}, a unified framework that brings this abstraction to life, facilitating efficient and scalable attention mechanism design across diverse hardware platforms.

\section{Design}
\label{sec:design}




\begin{figure}[t]
    \centering
    \includegraphics[width=0.7\linewidth]{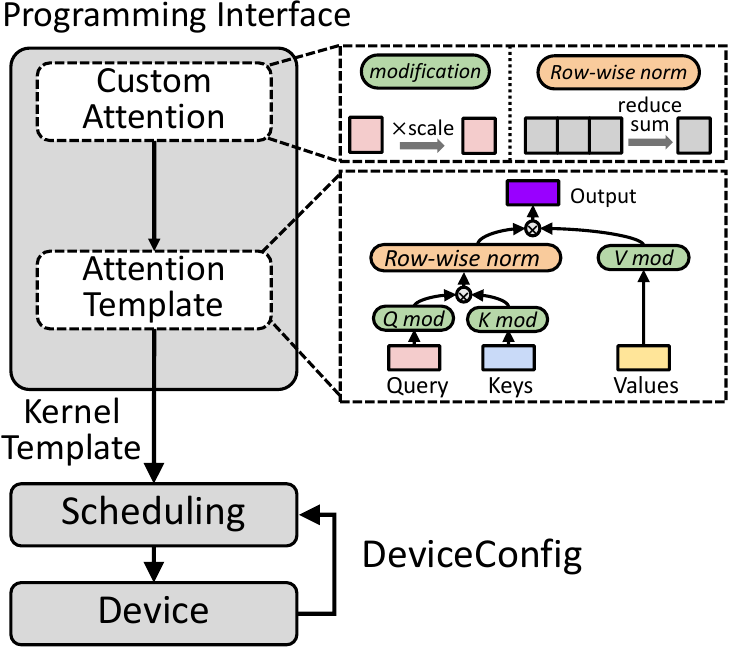}
    \caption{System overview: \oursys{} begins with attention templates in the Programming Interface to define Custom Attention. Then they are lowered to kernel templates and automatically scheduled to generate the best execution plan on the device.}
  \vspace{-5mm}
    \label{fig:overview}
\end{figure}

Expanding on our attention abstraction, we introduce \oursys{}, a unified framework designed to streamline the design, optimization, and execution of diverse attention mechanisms across hardware platforms. As shown in Figure~\ref{fig:overview}, \oursys{} begins with attention templates in the Programming Interface. These templates retain the core abstractions of attention—relevance scoring and aggregation—outlined in \S\ref{sec:attn} while providing customizable functions that allow users to design their own attention variants. By preserving the essential principles of attention and offering flexibility for user-defined extensions, \oursys{} facilitates the creation of a wide range of attention mechanisms and simplifies backend optimization.

Once customized, the attention mechanisms are lowered to \texttt{Kernel Templates}, which formalizes computation and memory operations. These templates, combined with two key components—\texttt{IntermediateTensor}, representing transient computational data, and \texttt{DeviceConfig}, capturing hardware constraints—define the scheduling space. \oursys{} employs a two-layer scheduling policy within this space to determine the optimal execution plan, balancing performance and resource utilization. The finalized plan is then mapped to hardware backends, ensuring scalability and efficiency across various configurations.

The following sections delve into the components of this framework, demonstrating how \oursys{} integrates abstraction, optimization, and execution to unify and extend the implementation of attention mechanisms.

\subsection{Programming Interface}




\begin{figure*}[t]
    \centering
    \includegraphics[width=0.9\linewidth]{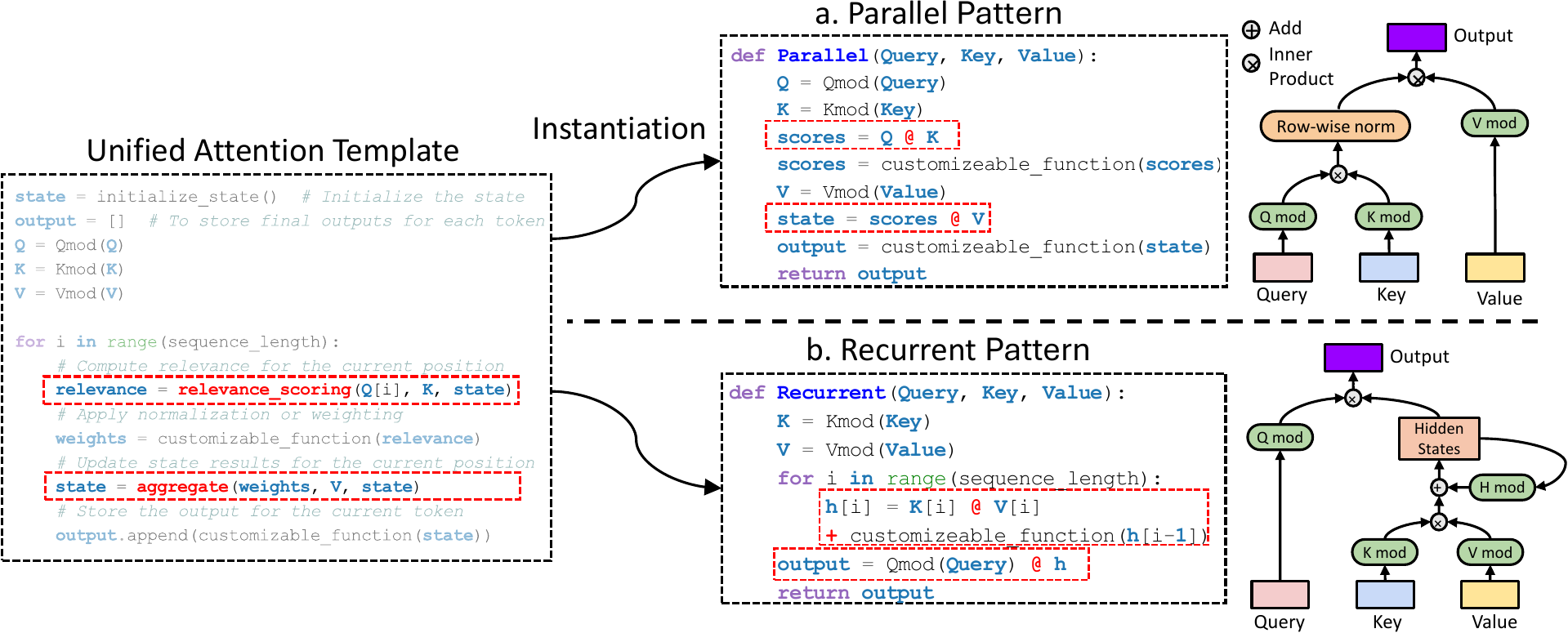}
  \vspace{-3mm}
    \caption{On the left is \oursys{}’s unified attention template. By instantiating this template, two distinct patterns are produced (Parallel Pattern and Recurrent Pattern). The red box highlights the operations corresponding to the core components of the attention mechanism in the unified attention template: \texttt{relevance\_scoring} and \texttt{aggregate}. Both the \texttt{customizable\_function} and the \texttt{mod} function are user-defined. The \texttt{customizable\_function} encompasses both \texttt{modification} function and \texttt{row-wise normalization} function, whereas the \texttt{mod} function is restricted to \texttt{modification} function only.}
  \vspace{-5mm}
    \label{fig:attn-template}
\end{figure*}





\begin{figure}
    \begin{lstlisting}
class modification:
    Func mod: Tensor->Tensor;
    Bool ismask;
class row-wiseNormalization:
    Func online_prologue: Tensor->Tensor;
    Func online_fwd: Tensor->Tensor;
    Func online_epilogue: Tensor->Tensor;
    \end{lstlisting}
  \vspace{-3mm}
    \caption{Customizable functions in programming interface}
  \vspace{-5mm}
    \label{fig:interface-pseudocode}
\end{figure}




\para{Attention Patterns and Templates}
Building on our abstraction of attention operations—relevance scoring and aggregation—we design a unified attention template that serves as a versatile foundation for implementing diverse attention mechanisms. As depicted in Figure~\ref{fig:attn-template}, the template takes $Q$, $K$, and $V$ after projection as inputs, retaining two fixed computations: $Q@K$ for relevance scoring and $S@V$ for aggregation. These computations capture the essence of attention mechanisms while offering flexibility through customizable functions.

The template includes two key customizable functions, \texttt{modification} and \texttt{row-wise normalization}, which can be inserted at designated points to enable users to define attention variants tailored to specific needs. These functions allow for operations such as applying masking, implementing custom normalization schemes, or adapting to unique computational goals.

To facilitate optimization, this unified template is instantiated in two computational patterns—parallel and recurrent:
\begin{itemize}[noitemsep,topsep=0pt, left=0pt]
    \item Parallel Pattern: Relevance scoring and aggregation are implemented as matrix multiplications, with $Q@K$ representing the scoring and $S@V$ representing the aggregation. Customizable functions are applied to the relevance scores to compute weights and to the state to produce the final output. Since most existing parallel attention variants do not innovate on state transformations, the customizable function for the state often defaults to an identity operation. This pattern is well-suited for mechanisms requiring global context and high parallelism.
    \item Recurrent Pattern: Relevance scoring and aggregation are sequentially computed, with $K@V$ and $Q@h$ together capturing the relevance scoring and aggregation, iteratively maintaining compressed states. In this pattern, the customizable functions on weights and states are reformulated as customizable function on the hidden state $h$. This makes the recurrent pattern ideal for memory-efficient designs and tasks with sequence dependencies.
\end{itemize}

By integrating the two instantiated patterns, this unified attention template empowers users to design high-level attention mechanisms while \oursys{} seamlessly handles low-level implementation and hardware-specific optimization, ensuring both efficiency and scalability.

\para{Customizable functions and flexibility.}
\label{subsec:func}
As shown in Figure~\ref{fig:interface-pseudocode}, customizable functions in \oursys{} include the \texttt{modification function} and the \texttt{row-wise normalization function}, which serve as user-defined components within the attention templates.

The \texttt{modification function} supports fine-grained elementwise transformations and masking, allowing users to customize operations applied to individual tensor elements. For example, scaling the query tensor by $1/\sqrt{d_k}$ in standard softmax attention can be achieved using this function. Masking operations, such as applying a causal mask, can also be implemented by annotating this function for masking purposes.

The \texttt{row-wise normalization function} provides a placeholder for normalizing or weighting. It enables global adjustments across tensor rows, accommodating a combination of elementwise and row-reduce computations. Examples include applying a row-wise softmax for normalizing attention scores or implementing numerical stabilization techniques. To enhance performance, the \texttt{row-wise normalization function} can be defined as an online function, where computations are processed sequentially in blocks along the rows. This approach significantly reduces memory overhead and ensures efficient execution.

\begin{figure}[t]
    \centering
    \includegraphics[width=1\linewidth]{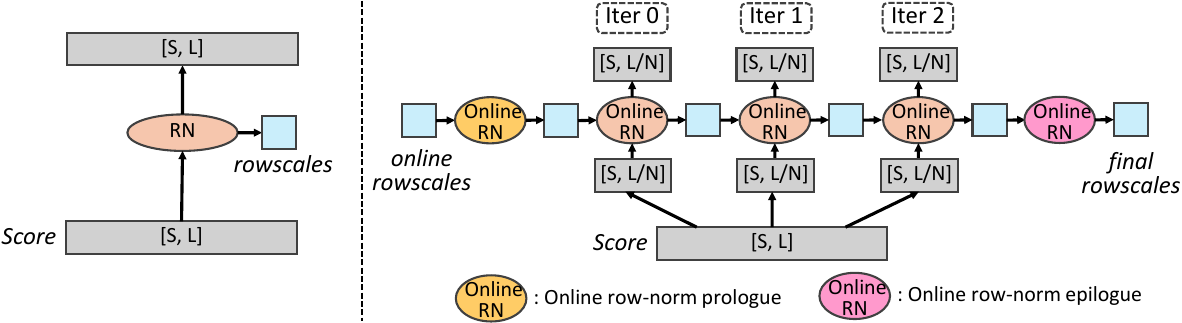}
    \caption{Illustration of the online row-norm interface. The left panel shows the standard row-wise normalization function, while the right panel demonstrates how \oursys{} enables users to implement the same functionality as an online function using the online row-norm interface.}
  \vspace{-5mm}
    \label{fig:onlineFunc}
\end{figure}

The \texttt{online row-norm interface} facilitates the implementation of online row-wise normalization, inspired by FlashAttention\cite{dao2022flashattention}. As shown in Figure~\ref{fig:onlineFunc}, this interface includes three main components:
\begin{itemize}[noitemsep,topsep=0pt]
    \item \texttt{online\_prologue}, which initializes the state variables before entering the online loop.
    \item \texttt{online\_fwd}, which defines computations within each block of rows, updating state variables like row maxima or sums.
    \item \texttt{online\_epilogue}, which finalizes the computation after the loop.
\end{itemize}

Users can leverage this interface to construct both forward and backward computation graphs via the \texttt{forward} and \texttt{backward} methods, enabling seamless integration with automatic differentiation and backend optimization. Key variables, such as \texttt{online\_rowscales} (state variables passed between blocks) and \texttt{final\_rowscales} (storing the final reduction results), provide the flexibility to define new online functions beyond softmax, significantly broadening the scope of attention mechanisms supported by \oursys{}.

\para{Computing primitives.}
To simplify user-defined operations, we introduce a set of \texttt{computing primitives}, which abstract hardware-specific details. These primitives are categorized as:
\begin{itemize}[noitemsep,topsep=0pt]
    \item Elementwise Operations: Operations like \textit{add(), sub(), tanh()}, and others allow fine-grained transformations of individual tensor elements.
    \item Row Reduce Operations: Aggregation functions such as \textit{reduceSum(), reduceMax()}, and \textit{reduceAbssum()} enable efficient row-wise reduce computations.
\end{itemize}
These primitives provide a robust foundation for defining both modification and row-wise normalization functions, ensuring compatibility with diverse hardware platforms.

\begin{figure}[t]
\begin{lstlisting}
def online_prologue():
    row_max = -inf
    row_sum = 0
def online_fwd(row_max_prev,row_sum_prev):
    row_max_cur = scores.reduceMax()
    row_max = max(row_max_cur, row_max_prev)
    
    scores = exp(scores-row_max)
    row_sum_cur = scores.reduceSum()
    row_sum = row_max_prev/row_max * row_sum_prev +
                row_sum_cur

    return row_max, row_sum
def online_epilogue():
    scores = scores / row_sum
\end{lstlisting}
  \vspace{-3mm}
\caption{Example of online row-wise normalization function: softmax attention.}
  \vspace{-5mm}
\label{fig:softmax-example}
\end{figure}

\para{Examples of Attention variants.}
Our interfaces support a wide range of attention mechanisms, demonstrating their flexibility and generality.

The Softmax Attention mechanism involves scaling the query tensor by $1/\sqrt{d_k}$ for normalization and applying a numerically stable softmax function to the scores. Specifically, the modification interface for $q$ is defined as $q\_mod = lambda\  q: q / \sqrt{d_k}$, while the row-wise normalization interface on scores is expressed as: $score\_rownorm = lambda\ scores: (\_scores = \exp(scores - scores.\text{reduceMax()}); \_scores = \_scores / \_scores.\text{reduceSum()})$. To enhance performance, we implement the row-wise normalization function in an online form using our online row-norm interface (Figure~\ref{fig:softmax-example}). Specifically, \texttt{online\_prologue} initializes the state, \texttt{online\_fwd} performs intermediate computations on scores and state, and \texttt{online\_epilogue} finalizes the computation.

ReluAttention replaces the softmax function with a row-wise normalization function that contains only an elementwise operation. We use our modification interface on scores as $score\_mod = lambda\ scores: \max(scores, 0)$. In this case, no additional normalization is applied.

Similarly, in RetNet parallel attention, a retention mask is applied to the scores. This is represented by the modification function $ score\_mod = lambda:\ scores = scores \times mask$, and a row-wise normalization function ensures numerical stability, defined as $score\_rownorm = lambda\ scores:  (scores / scores.\text{abs()}.\text{reduceSum()}.\text{clamp}(\text{min}=1))$.

Mamba2, a representative linear attention mechanism, incorporates a selective gating mechanism to modulate the key and hidden states, allowing selective attention to past information. Our interface represents this as a modification function on the key $k\_mod = lambda\ k: k \times gate$ and a modification function on the hidden states $h\_mod = lambda\ h: h \times decay \times gate$.

\subsection{Scheduling Space}
The scheduling space in \oursys{} is inherently shaped by the kernel templates, which encapsulate the computation flow of attention mechanisms. These templates, derived from our attention pattern abstractions, constraining the range of scheduling options while enabling efficient and adaptable execution. Together with \texttt{IntermediateTensor} and \texttt{DeviceConfig} components, the kernel templates form the foundation for determining optimal execution strategies.

\para{Kernel template.}
Kernel templates play a pivotal role in structuring the scheduling space by formalizing the computation and memory operations of attention mechanisms. These templates provide a consistent structure for implementing diverse attention mechanisms while allowing flexibility for hardware-specific optimizations. For example, templates designed for parallel patterns incorporate online techniques to efficiently manage row-wise normalization, while those for recurrent patterns utilize chunk parallelism to maximize tensor core utilization and computational efficiency.

Additionally, \oursys{} supports multiple kernel templates tailored to different hardware backends, such as those implemented in Triton\cite{triton}, CUTE\cite{nvidia2024cutlass}, and TileLang\cite{tilelang}. Leveraging a common lowering method based on attention pattern abstractions, \oursys{} ensures that customized attention variants can be seamlessly lowered to these templates. This flexibility allows \oursys{} to dynamically select the optimal kernel template based on the input data and hardware platform, achieving consistent high performance across configurations.

\begin{figure}[t]
\begin{lstlisting}
class IntermediateTensor{
    TileShape tile;
    MemoryLocation mem;
    int pipelineStage;
};
\end{lstlisting}
  \vspace{-3mm}
\caption{IntermediateTensor component}
  \vspace{-5mm}
\label{fig:intermediate-tensor}
\end{figure}

\begin{figure}[t]
\begin{lstlisting}
class DeviceConfig{
    BaseTileShape basetile;
    List<MemoryCapacity> memoryInfo;
};
\end{lstlisting}
  \vspace{-3mm}
\caption{DeviceConfig component}
  \vspace{-5mm}
\label{fig:device-config}
\end{figure}

\para{IntermediateTensor.}
At the heart of the scheduling space lies the \texttt{IntermediateTensor} component, which encapsulates the transient data generated during computation. By focusing on intermediate tensors, \oursys{} can systematically deduce the tiling, memory allocation, and pipeline requirements for attention mechanisms.

Key attributes of \texttt{IntermediateTensor} include:
\begin{itemize}[noitemsep,topsep=0pt]
    \item Tensor tile shape (\texttt{tile}): By dividing tensors into smaller tiles, we can perform operations tile-by-tile and allocate buffers efficiently. Using the computation graph, we propagate the tiling scheme across all operations to infer the tile shapes of $Q$, $K$, $V$ and other tensors, ensuring an optimal balance between computation and memory.
    \item Tensor location (\texttt{mem}): Intermediate tensors can be stored in various levels of memory, such as global memory, shared memory, or registers. Each location offers a trade-off between latency, bandwidth, and resource availability. 
    \item Pipeline stage (\texttt{pipelineStage}): Operations involving intermediate tensors are divided into multiple pipeline stages, such as memory copy and computation. The number of stages determines the buffer requirements and scheduling flexibility, enabling overlapping operations to maximize throughput and minimize resource contention.
\end{itemize}
This component ensures that all elements of the attention mechanism, including inputs, outputs, and intermediate results, are unified under a consistent scheduling strategy.

\para{DeviceConfig.}
The \texttt{DeviceConfig} component provides hardware-specific constraints that refine the scheduling space defined by kernel templates and intermediate tensors. It encapsulates attributes such as:
\begin{itemize}[noitemsep,topsep=0pt]
    \item Base tile shape (\texttt{basetile}): Specifies the optimal tile shape for computations on the target hardware, ensuring alignment with hardware-specific constraints, such as alignment with GEMM computing instruction and memory transaction.
    \item Memory hierarchy (\texttt{memoryInfo}): Provides details about the available memory tiers (e.g., registers, shared memory, global memory) and their respective capacities, enabling efficient allocation and minimizing contention.

\end{itemize}

\texttt{DeviceConfig} plays a pivotal role in determining the feasible tiling and memory strategies during scheduling. For instance, the base tile shape ensures hardware-aligned tiling configurations, while memory capacity constraints prevent resource overcommitment.

By combining kernel templates, \texttt{IntermediateTensor}, and \texttt{DeviceConfig}, \oursys{} constructs a unified scheduling space that supports diverse attention mechanisms and hardware platforms. Kernel templates anchor the computation flow, \texttt{IntermediateTensor} defines the key computational attributes, and \texttt{DeviceConfig} introduces hardware constraints, together forming a robust and scalable scheduling framework.

\subsection{Scheduling policy}

As illustrated in Figure~\ref{alg:schedule}, \oursys{} employs a two-layer scheduling policy to minimize latency and optimize execution. This policy operates at two levels: \texttt{tile config scheduling} and \texttt{tile resource scheduling}. At the tile config scheduling level, the policy traverses the entire space of possible tile configurations, leveraging the constrained nature of the scheduling space to perform exhaustive exploration. At the tile resource scheduling level, the policy determines the optimal memory placement and execution strategy within each tile configuration, ensuring efficient hardware resource utilization while adhering to hardware constraints.

\begin{algorithm}[t]
    \SetKwProg{Fn}{Func}{}{end}
    \SetKwIF{If}{ElseIf}{Else}{if}{}{else if}{else}{end if}
    \Fn{TileConfigScheduling(g: Graph, D:DeviceConfig)}
    {
    tensor\_tile\_configs = InferPossibleTileConfigs(g, D.basetile);
    \label{alg:infer_tile}

    plans = []
    
    \For{tile\_config in tensor\_tile\_configs}{
        \label{alg:traverse_tileconfig}
        plans += TileResourceScheduling(tile\_config, g.IntermediateTensors , D);
        \label{alg:intra_tile}
    }
    \For{plan in plans}{
    \label{alg:traverseplans}
        \If{Profile(plan) < best\_latency}{
        \label{alg:profile}
          best\_latency = Profile(plan);
          best\_plan = plan;
        }
    }
    return best\_plan;
    }
    \Fn{TileResourceScheduling(tile\_config: TileConfig, t: IntermediateTensors, D:DeviceConfig)}
    {
        InitMemLocation(t.memLoc, REGISTER);
        \label{alg:initmemloc}
        
        t = sortByTensorSizeDec(t);
        \label{alg:sorttensor}

        \For{tensor\_i in t}{
        \label{alg:iter_tensor}
        
        plans = GeneratePlans(t);
        \label{alg:generateplans}
        
        \For{plan in plans}{
            \If{not ComputeMemoryConstraint(tile\_config, t, plan, D.memoryInfo)}{
            \label{alg:computememconstrain}
                plans.remove(plan);
            }
        }

        \If{not plans.isEmpty()}{
        \label{alg:plan_isempty}
            return plans;
        }

        LowerMemLocation(tensor\_i.memLoc)
        \label{alg:lowermemloc}
        } 
    return EmptySet();
    \label{alg:return}
    }
\caption{Scheduling algorithm}
\label{alg:schedule}
\end{algorithm}

\para{Tile config scheduling.} 
The tile config scheduling layer takes as input a computation graph (\texttt{Graph}) composed of \texttt{IntermediateTensor} objects and hardware configuration details (\texttt{DeviceConfig}). This layer begins by invoking the \texttt{InferPossibleTileConfigs} function (line \ref{alg:infer_tile}) to identify all potential tile configurations for the computation graph, propagating from the output tensors. Due to the complexity of attention mechanisms, including their intricate computation stages, hardware alignment requirements, and memory limitations, the tile configuration space is constrained. This enables an exhaustive traversal of all possible tile configurations.

For each tile configuration (line \ref{alg:traverse_tileconfig} - \ref{alg:intra_tile}), the policy generates a set of execution plans using the tile resource scheduling layer and evaluates their performance through profiling (line \ref{alg:traverseplans} -\ref{alg:profile}). Profiling involves calculating the latency of each plan to determine its efficiency. Finally, the tile configuration corresponding to the plan with the lowest latency is selected as the optimal configuration. 

\para{Tile resource scheduling.} 
The tile resource scheduling layer optimizes the execution plan for a specific tile configuration. The process starts by initializing all intermediate tensors to the highest memory tier available (e.g., registers) to reduce memory I/O overhead (line \ref{alg:initmemloc}). The intermediate tensors are sorted in descending order of size, prioritizing larger tensors for memory allocation to maximize efficiency (line \ref{alg:sorttensor}).

For each tensor, the policy iteratively generates execution plans (line \ref{alg:iter_tensor}-\ref{alg:return}) and checks their feasibility against hardware constraints, such as memory capacity and alignment requirements (line \ref{alg:computememconstrain}). If no valid plan is found, the policy progressively demotes tensors to lower memory tiers (e.g., shared or global memory) and reattempts plan generation (line \ref{alg:plan_isempty} -\ref{alg:lowermemloc}). This iterative adjustment continues until a feasible plan is identified or all options are exhausted. If no valid plan can be generated, the function returns an empty set (line \ref{alg:return}).

By combining the two layers, the scheduling policy systematically explores the design space to produce efficient, hardware-aware execution plans for attention computations. This hierarchical approach enables \oursys{} to balance performance and resource utilization, supporting diverse attention variants across multiple hardware platforms.

\section{Implementation}
\label{sec:impl}

\begin{figure}
    \centering
    \includegraphics[width=0.7\linewidth]{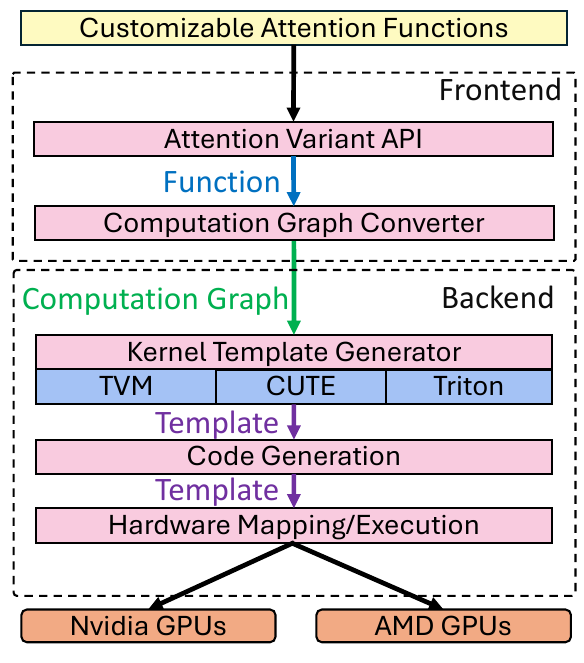}
  \vspace{-3mm}
    \caption{Implementation of \oursys{}}
    \vspace{-5mm}
    \label{fig:implementation}
\end{figure}

In this section, we present the implementation of the \oursys{} frontend and backend. Figure~\ref{fig:implementation} provides an overview of the \oursys{} workflow. We intergrate \oursys{} into pytorch\cite{pytorch} as a module. The frontend accepts user-defined functions as input, constructs the computation graph of intermediate tensors by graph tracer, and passes it to the backend, which generates optimized device kernels for efficient execution.

\subsection{Frontend}
\label{sec:frontend}

The frontend of \oursys{} provides the foundation for defining and representing user-defined attention mechanisms. It introduces a set of computing primitives, facilitates the use of customizable functions such as modification and row-wise normalization. Furthermore, \oursys{} traces computation graphs by encoding tensor attributes, enabling automatic differentiation and efficient backend integration.

\para{Computing primitives.}
The frontend includes a rich set of computing primitives, categorized into elementwise and row-reduce operations. Elementwise operations, such as \textit{add(), sub(), tanh()}, are computed in a SIMT style on GPUs and are fused with matrix multiplications at the register or shared memory level to minimize memory access overhead. Row-reduce operations, such as \textit{reduceSum()} and \textit{reduceMax()}, leverage GPU warp-level reduction, where each row-reduce operation is computed by the same thread block and warp. 

\para{Modifaction function.} 
The modification function exclusively supports elementwise operations. These operations are fused by \oursys{} into a single computation unit and lowered to the backend kernel template for efficient execution. The modification function also supports masking operations, allowing users to implement attention variants that require masking logic.

\para{Row-wise normalization.}
The row-wise normalization function supports both elementwise and row-reduce operations or a combination of the two. Similar to the modification function, all operations within the row-wise normalization function are fused and lowered to backend kernel templates.

\para{Tracing user-defined computation graphs.}
\oursys{} traces user-defined computation graphs by building a directed acyclic graph of Tensor. Each node contains the computing primitive(such as $add()$, $reduceSum()$), the coresponding output Tensor attributes(such as shape), and a list of pointers pointing to its preceded node. This enables the system to dynamically trace the dependency between tensors.
We also define a $grad$ field on each node, which is a pointer to another node containing the gradient of current tensor. By iteratively traverse between nodes, we encodes the gradients informantion of each node into the $grad$ field to achieve automatic differentiation.
The forward and backward computing graph \oursys{} constructed ensures seamless integration with the backend for efficient kernel generation.


\subsection{Backend}
\label{sec:backend}
The backend of \oursys{} transforms user-defined algorithms into kernel templates and optimizes these templates into high-performance kernels.

\para{Kernel template.}
We design kernel templates to systematically implement attention lowering. Since most custom attention mechanisms preserve the overall kernel computation flow, a template-based approach is particularly effective. Using the computation graphs generated from the \texttt{modification} and \texttt{row-wise normalization} functions in \S\ref{sec:frontend}, we produce essential components such as intermediate tensor definitions, initialization routines, memory operations, and computation steps. These components are seamlessly fused into the kernel templates, ensuring computational and memory efficiency.

To achieve extreme performance, we leverage TileLang\cite{tilelang} and CUTE\cite{nvidia2024cutlass} to implement optimized kernel templates. For parallel attention, our templates employ advanced online techniques to handle row-wise normalization efficiently, ensuring adaptability across a wide range of configurations. For recurrent attention, we utilize chunk parallelism to fully exploit tensor cores, balancing computational throughput and efficiency. These strategies allow \oursys{} to accommodate the unique characteristics of different attention mechanisms while maximizing hardware utilization.

Handcrafted kernels, such as FlashAttention, are often limited to specific configurations, typically requiring \texttt{$d_{qk}$} to equal \texttt{$d_v$}. This highlights a key shortcoming of ad hoc approaches: they fail to address diverse input configurations without extensive manual tuning of schedules, such as tile sizes, pipelining, and fusion strategies—efforts that demand significant expertise. In contrast, \oursys{} automates this process, enabling support for various input configurations without manual intervention. Our kernel templates are designed to handle diverse configurations of \texttt{$d_{qk}$} and \texttt{$d_v$}, eliminating the need for padding when these dimensions differ, as seen in models like DeepSeek V2 (\texttt{$d_{qk}$}=192, \texttt{$d_v$}=128)\cite{deepseekai2024deepseekv2strongeconomicalefficient} and DiffTransformer (\texttt{$d_{qk}$}=128, \texttt{$d_v$}=256)\cite{ye2024differentialtransformer}. By reducing padding overhead and computation costs, \oursys{} not only extends support to a broader range of attention designs but also enhances performance while maintaining flexibility across hardware platforms.

\para{Lowering computation graphs to kernel templates.} The lowering process translates user-defined computation graphs into kernel templates. This process is divided into two stages: expression generation and code generation. The split design enhances extensibility, with expression generation being kernel-template-agnostic and code generation adapting to specific kernel templates.

During expression generation, \oursys{} inputs a user-defined computation graph and performs a topological sort to convert it into a linear sequence of computation expressions, preserving the computation order. Additionally, as the graph is traversed, the use-define chain for each node is analyzed, enabling optimizations such as variable reuse. In the subsequent code generation phase, these computation expressions are used to produce kernel code tailored to the selected kernel template through string matching. The resulting kernel code includes variable initialization, memory copying, and computation steps, seamlessly integrating user-defined operations into efficient kernel templates.






\para{Map to hardware backend.}
We map the kernel templates to both NVIDIA GPUs and AMD GPUs, optimizing performance across diverse hardware platforms.

For NVIDIA GPUs, \oursys{} supports two backends: TileLang\cite{tilelang} and CUTE\cite{nvidia2024cutlass}. Using the Triton-like compiler, we map elementwise operations and reduce operations by utilizing APIs such as \texttt{ParallelFor} for thread-level execution and \texttt{reduce\_sum}/\texttt{reduce\_max} for block-level row-reduction. With CUTE, we employ \texttt{cute::Tensor} and \texttt{cute::layout} to define thread-level data layouts and map reduce operations to efficient micro-kernel templates, ensuring high performance for compute-intensive tasks.

For AMD GPUs, \oursys{} supports the MI250, AMD’s high-performance GPU architecture, equipped with Matrix Cores for matrix multiplication, Arithmetic Logic Units (ALUs), and asynchronous copy units for efficient memory transfer. Leveraging TileLang\cite{tilelang}’s capabilities, we generate highly optimized kernels tailored to the MI250, fully utilizing its advanced hardware features for efficient execution.

\section{Evaluation}
\label{sec:eval}

\begin{table}[t]
\centering
\resizebox{0.48\textwidth}{!}{
\begin{tabular}{c|c|c}
\hline
\textbf{Operator} & \textbf{Configuration} & \textbf{Model} \\ \hline
\multicolumn{1}{c|}{\textbf{}} & \multicolumn{1}{c|}{} & \multicolumn{1}{c}{} \\[-0.9em]  \hline
Softmax Attention & head=32, dimqk=128, dimv=128 & LLAMA3.1-8B \\
Softmax Attention & head=16, dimqk=192, dimv=128 & Deepseek-V2-lite   \\ 
Softmax Attention & head=12, dimqk=128, dimv=256 & DiffTransformer-3B   \\ \hline
Sigmoid Attention & head=32, dimqk=128, dimv=128 & LLAMA3-8B-style  \\ \hline
Relu Attention & head=6, dimqk=64, dimv=64 & ViT-s/16-style   \\ \hline
Retention Parallel & head=32, dimqk=256,dimv=512 & RetNet-6.7B   \\\hline
Mamba2 SSM & headv=80, dimqk=128, dimv=64 & Mamba2-2.7B   \\ \hline
Retention Recurrent & head=32, dimqk=256,dimv=512 & RetNet-6.7B  \\ \hline
Gated Retention & head=40, dimqk=256, dimv=256 & YOCO-13B   \\ 
Gated Retention & head=16, dimqk=64, dimv=64 & RFA-Big   \\ \hline
Softmax Attention Decoding & Seqlen=1, head=16, dimqk=192, dimv=128 & Deepseek-V2-lite  \\ \hline
\end{tabular}
}
\caption{A subset of attention in our microbenchmark.}
\label{table:operators}\vspace{-5mm}
\end{table}

\begin{figure*}[t]
    \centering
    \includegraphics[width=0.98\linewidth]{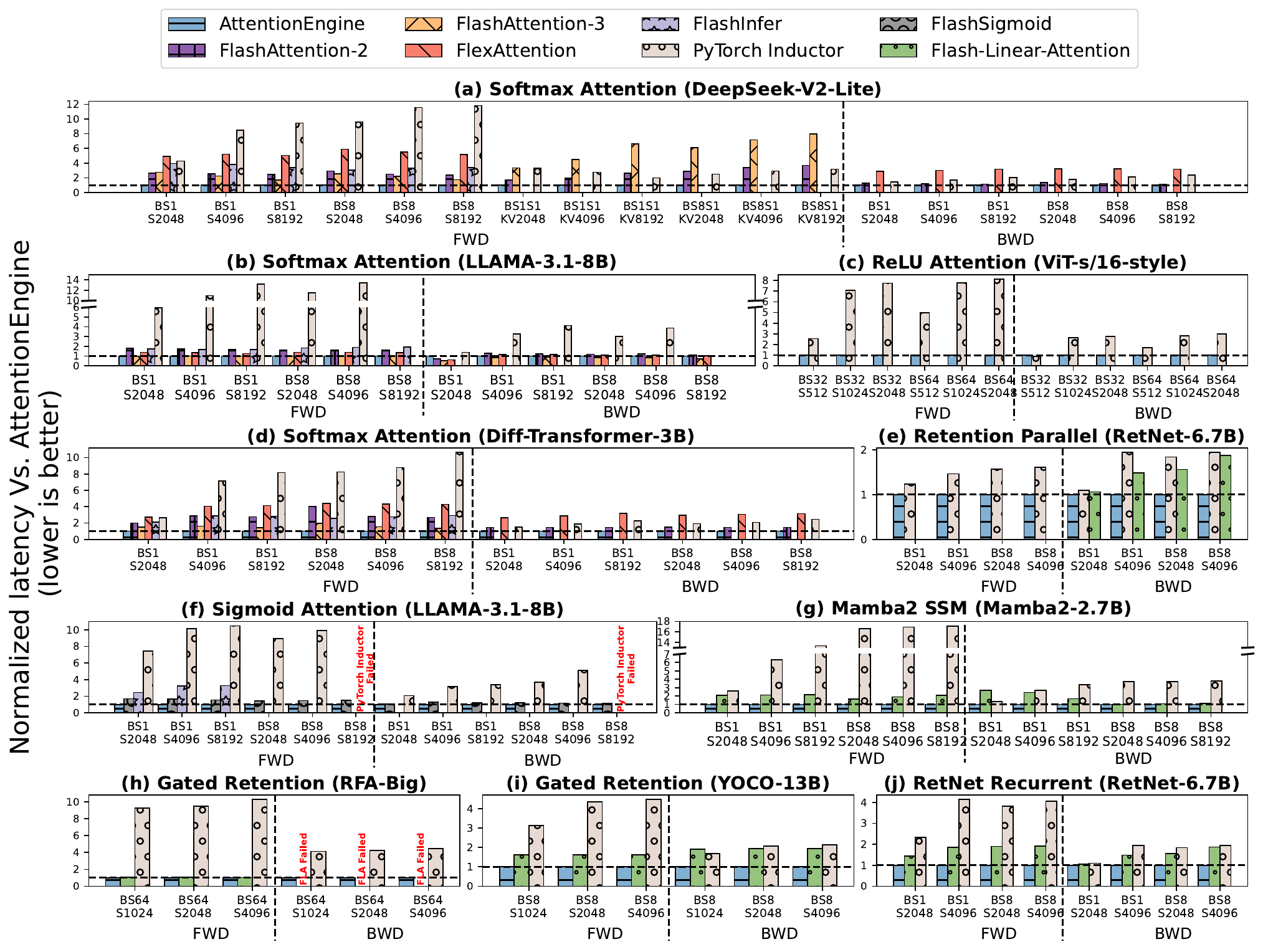}
    \caption{Attention operator performance on H100 GPUs.}
    \vspace{-3mm}
    \label{fig:eval-fwd}
\end{figure*}

In this section, we evaluate \oursys{} on both attention microbenchmarks and end-to-end models by comparing them with the
state-of-the-art libraries and the compiler-based method to demonstrate the effectiveness of \oursys{}.
We summarize our findings: (1) \oursys{} can optimize standard transformer attention, achieving comparable performance with hand-crafted libraries. (2) \oursys{} can generate custom attention kernels, achieving speedup up to $10.4\times$. (3) \oursys{} support multi-backends, including NVIDIA and AMD GPUs.

\subsection{Experimental Setup}

\paragraph{Hardware platforms.} We evaluate \oursys{} on both NVIDIA and AMD GPUs, as they are currently the most popular hardware platforms. Our evaluation includes two high-performance GPUs: the NVIDIA H100 and the AMD Instinct MI250 GPU. 
We use CUDA version 12.4, Triton version 2.3.1 with the H100 GPU, and the ROCm version 6.2.4, Triton version 3.1.0 with the MI250 GPU. Both GPUs are evaluated on the operating system Ubuntu 20.04.


\para{Attention workload.} We evaluate eight Attention algorithm, including four parallel pattern attention (Softmax Attention\cite{Ashish17AttentionIsAllYouNeed}, Sigmoid Attention\cite{ramapuram2024sigmoidattn}, ReLU Attention\cite{wortsman2023replacingsoftmaxreluvision} and parallel form of multi-scale retention\cite{sun2023retentive}) and four recurrent pattern attention (mamba2\cite{dao2024mamba2}, random feature attention\cite{peng2021randomfeatureattention}, retention rucurrent\cite{sun2023retentive}, gated retention\cite{sun2024YOCO}) 
For softmax attention, we perform the tests using configuration of LLAMA3.1-8B\cite{dubey2024llama3}, Deepseek-V2-lite\cite{deepseekai2024deepseekv2strongeconomicalefficient} and DiffTransformers-3B\cite{ye2024differentialtransformer}.
We select the batch size as 1 and 8 and sequence length as 2k, 4k and 8k for attention in large language models, which are common configurations for these models. 
Table \ref{table:operators} lists a representative subset of operators as well as
their configurations.

\para{Baselines.} We compare \oursys{} with manually implemented attention libraries, such as FlashAttention-v2\cite{dao2023flashattention} and FlashAttention-v3 \cite{shah2024flashattention} for Softmax attention, FlashSigmoid\cite{ramapuram2024sigmoidattn} for Sigmoid attention, Mamba2 chunk kernel\cite{dao2024mamba2} for Mamba2 SSM and Flash-Linear-Attention triton library\cite{yang2024fla} for gated retention. 
We also compare with state-of-the-art programming model-based approaches, such as FlexAttention\cite{dong2024flexattentionprogrammingmodel} and FlashInfer\cite{ye2025flashinferefficientcustomizableattention} for transformer attention. We use PyTorch\cite{pytorch} as a default baseline for attention that does not have a manually-implemented library, such as Retention Parallel\cite{sun2023retentive} and ReLUAttention\cite{wortsman2023replacingsoftmaxreluvision}.

\subsection{Attention Performance on NVIDIA H100}

Figure \ref{fig:eval-fwd} shows the performance of attention performance on NVIDIA H100. The x-axis represents different configs of attention operators, and the y-axis indicates the normalized latency relative to \oursys{}.

\para{Softmax attention.}
Figure \ref{fig:eval-fwd} (a)(b)(d) shows the performance of \oursys{} and other baselines on Softmax attention from Deepseek-V2-Lite, LLAMA3.1-8B, and Diff-Transformer-3B.
Compared with highly optimized libraries, \oursys{} still obtain significant speedup because of more flexible kernel templates. Compared with highly-optimized FlashAttention, \oursys{} achieves an average speedup of $1.88\times$ for forward and $1.52\times$ for backward on DeepSeek-V2-Lite and Diff-Transformers-3B, and achieves comparable performance on LLAMA3.1-8B. This improvement stems from \oursys{}'s flexible kernel template to natively support different \texttt{headdim\_qk} and \texttt{headdim\_v}, instead of padding them to the same dimension. \oursys{} also outperforms other programming-model-based approaches such as FlexAttention and FlashInfer, due to our scheduling over different shapes.

\para{Customized transformer attention.}
Figure \ref{fig:eval-fwd} (c)(e)(f) shows the performance of \oursys{} and other baselines on 
customized transformer attention (Sigmoid attention, ReLU attention and retention parallel). Current expert-optimized libraries lack support for these custom attentions. For example, no fused attention kernel is implemented for ReLU attention and fused Sigmoid attention kernel is not optimized for the latest hardware like NVIDIA H100. \oursys{} can obtain significant speedup on these customized attentions, achieving $3.6\times$ ($1.1\times\sim 10.4\times$) over FlashSigmoid, PyTorch ReLU attention and PyTorch retention parallel. In addition, compared with programming-model-based approaches, \oursys{} can support all three customized attention, which demonstrates \oursys{}'s expressive ability and scalability.

\para{Mamba2.}
Figure \ref{fig:eval-fwd} (g) represents the linear attention operation of the Mamba2 model: State Space Module. We compare \oursys{} with the official Mamba2 implementation using Triton. 
\oursys{} achieves average speedups of $1.99\times$ and $1.65\times$ over Triton for Mamba2 forward and Mamba2 backward, respectively. This demonstrates the complexity of manually optimizing the attention kernel and the necessity of \oursys{}.

\para{Retention and gated retention.}
Figure \ref{fig:eval-fwd} (h)(i)(j) represents the linear attention operation of RetNet-6.7B, YOCO-13B and RFA-Big.We compare \oursys{} with Flash-Linear-Attention, which is an expert-optimized linear attention library. The result show that \oursys{} achieves average speedups of $1.33\times$ and $1.93\times$ for forward and backward, respectively.

\subsection{End-to-end Inference on NVIDIA H100}

\begin{figure}[t]
    \centering
    \includegraphics[width=0.85\linewidth]{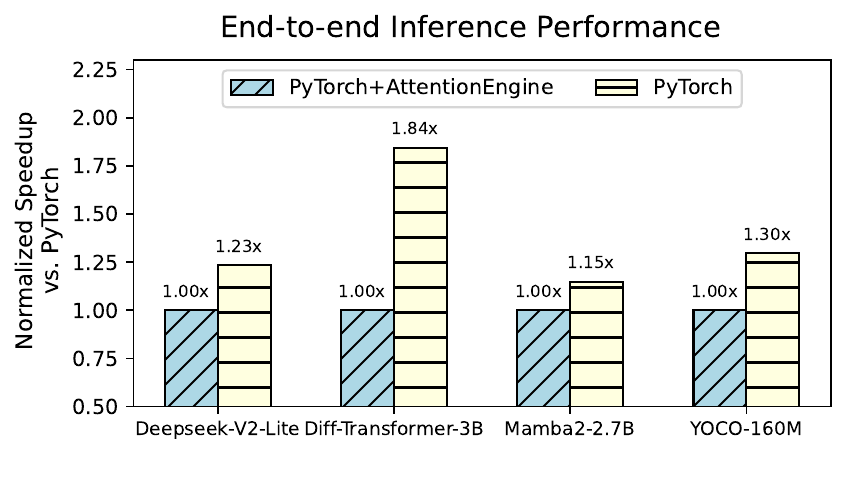}
    \vspace{-5mm}
    \caption{End-to-end inference performance on H100.}
    \vspace{-5mm}
    \label{fig:end-to-end inference}
\end{figure}

We evaluate the inference latency of large language models like DeepSeek-V2-Lite and Mamba2-2.7B. We show \oursys{}'s applicability to end-to-end inference.

\para{Inference setup.} We evaluate end-to-end inference on one NVIDIA H100 GPU. We use Transformers\cite{wolf-etal-2020-transformers} for end-to-end inference, which is the most popular machine learning framework and is backed by PyTorch. We test two models with parallel pattern attention (Deepseek-V2-Lite and Diff-Transformer-3B ) and two models with recurrent pattern attention (Mamba2-2.7B and YOCO-160M).  We replace the attention operator in these models with \oursys{}.

\para{Inference performance.} As shown in Figure \ref{fig:end-to-end inference}, \oursys{} acheive an average speedup of $1.4\times$ on these models with FP16 precision. These speedup came from our more efficient attention operator. For example, In DeepSpeed-V2-Lite, attention accounts for $85\%$ of the total inference time. We improved the attention operator's speed to $2.2\times$ by supporting different head dimensions for \texttt{q}, \texttt{k}, and \texttt{v}, thereby enhancing the end-to-end performance to $1.85\times$.

\subsection{End-to-end Training on NVIDIA H100}

\begin{figure}[t]
    \centering
    \includegraphics[width=0.85\linewidth]{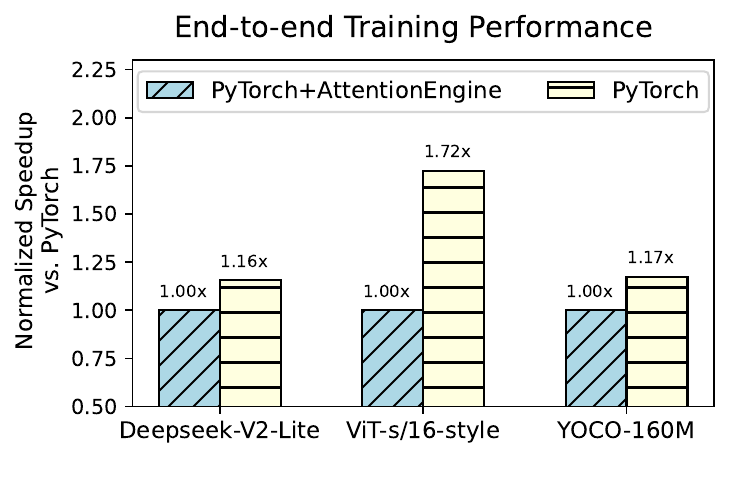}
    \vspace{-5mm}
    \caption{End-to-end training performance on H100.}
    \vspace{-5mm}
    \label{fig:end-to-end-train}
\end{figure}

We also evaluate end-to-end training of attention-based model and linear attention-based model to demonstrate \oursys{}'s ability in both forward and backward.

\para{Training setup.} We use TRL\cite{vonwerra2022trl} for training, which is a full stack library based on transformers that provides a set of tools to train transformer language models. Our workloads are Diff-Transformer-3B, YOCO-160M and ViT-S/16 with ReLU attention.

\para{Training performance.} As shown in Figure \ref{fig:end-to-end-train}, we achieve an average speedup of $1.4\times$ on these models. For ViT-S/16 with ReLU attention, we achieve $1.7\times$ speedup due to the lack of existing libraries for ReLU attention.

\vspace{-2mm}
\subsection{Evaluation on AMD ROCm GPUs}
\vspace{-2mm}

\begin{figure}
    \centering
    \includegraphics[width=0.9\linewidth]{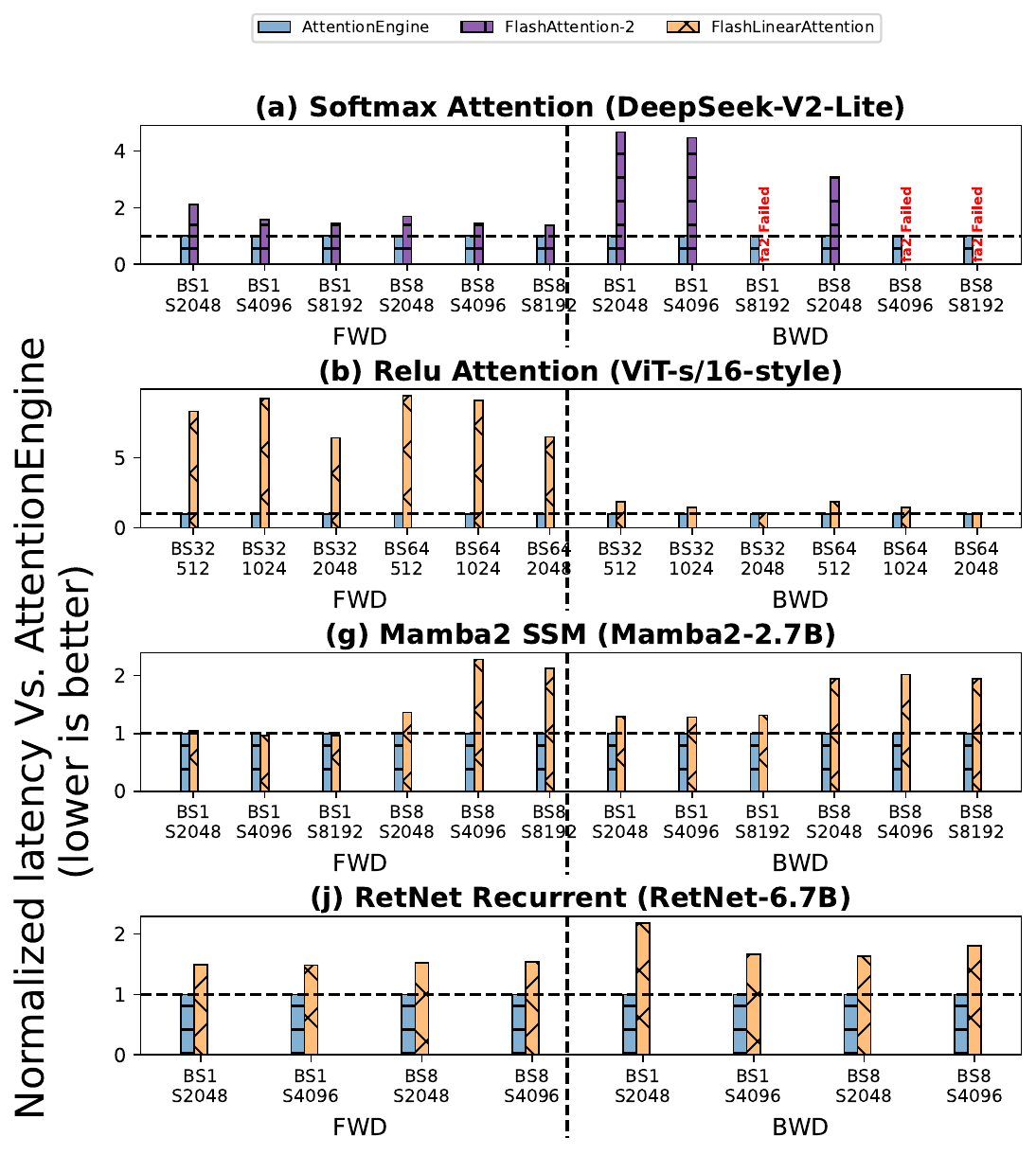}
    \vspace{-5mm}
    \caption{Attention operator performance on MI250 GPUs.}
    \vspace{-5mm}
    \label{fig:mi250-eval}
\end{figure}

We benchmark the AMD MI250 GPU using a subset of operators selected from the microbenchmark suite originally designed for the NVIDIA H100 GPU, including Softmax Attention, ReLU Attention, Mamba2 and RetNet Recurrent.

Figure \ref{fig:mi250-eval} shows that \oursys{} outperforms an average of $3.3\times$ for forward and $2.0\times$ for backward over other baselines across different attention operators. 
This demonstrated \oursys{}'s ability to support multi-backend.

\vspace{-2mm}
\section{Related Work}
\vspace{-2mm}

\para{Handcrafted attention.}
High-performance attention mechanisms frequently rely on handcrafted kernel implementations optimized for specific patterns. FlashAttention\cite{dao2022flashattention} provides a highly optimized kernel for standard transformer attention, utilizing techniques such as online softmax, memory-efficient fusion, and pipelining. It is implemented using CUTE\cite{nvidia2024cutlass} on NVIDIA GPUs and ComposableKernel on AMD GPUs for low-level optimization.  Mamba2\cite{dao2024mamba2}, with official kernels developed in Triton\cite{triton}, focuses on tensor core utilization to enhance efficiency. Flash-Linear-Attention\cite{yang2024fla}, a third-party repository, extends beyond individual methods like Mamba2 and Gated Linear Attention (GLA), offering kernels for a wide variety of linear attention variants.

FlexAttention\cite{dong2024flexattentionprogrammingmodel} and FlashInfer\cite{ye2025flashinferefficientcustomizableattention} aim to simplify the development of attention mechanisms by offering high-level abstractions. However, these approaches primarily focus on elementwise transformations within transformer attention and are exclusively targeted at NVIDIA GPUs. While effective in their domain, their scope is limited, excluding support for linear attention and more advanced optimization strategies. Additionally, their lack of compatibility with AMD GPUs highlights a significant gap in addressing multi-backend requirements.

While these implementations achieve excellent performance, they are restricted to specific attention designs and require substantial manual effort to adapt for new variants. This reliance on handcrafted kernels limits scalability and slows innovation, particularly for emerging attention designs. In contrast, \oursys{} abstracts the complexity of kernel development, enabling users to define and optimize diverse attention mechanisms without the need for manual implementation. By leveraging a unified programming model and automated optimization pipeline, \oursys{} supports a broader range of configurations while maintaining competitive performance.

\para{Compiler optimization.}
Existing DNN compilers, such as TVM\cite{tvm2018}, Ansor\cite{ansor}, XLA\cite{xla}, Welder\cite{shi2023welder}, Ladder\cite{wang2024ladder}, and TensorRT\cite{tensorrt}, widely adopt techniques like operator fusion to reduce memory overhead and improve computational efficiency. However, these approaches primarily focus on spatial tiling for regular operators, neglecting the unique challenges and opportunities presented by attention mechanisms. \oursys{} incorporates common compiler optimization methods, such as fusion and tiling, while extending them to support the irregular computations inherent in attention mechanisms. 


\oursys{} overcomes these limitations by supporting both transformer and linear attention within a single framework. It incorporates advanced scheduling techniques and targets multiple backends, including NVIDIA and AMD GPUs, ensuring high performance and scalability. By unifying diverse attention mechanisms under a comprehensive programming model, \oursys{} facilitates the efficient development and deployment of a wide range of attention designs across heterogeneous hardware architectures.

\section{Conclusion}
\label{sec:conclusion}

Attention mechanisms are central to transformers and large language models (LLMs), driving advancements in natural language processing by capturing contextual relationships. However, their computational demands and growing design diversity pose challenges for scalability and optimization. \oursys{} addresses these issues by abstracting attention into two core operations, i.e., relevance scoring and aggregation, and introducing customizable templates that combine flexibility with efficiency. With a cross-backend scheduling framework, \oursys{} automates kernel optimizations, achieving up to 10.4$\times$ speedups for unsupported configurations and providing a foundation for diverse attention designs.
\clearpage

\bibliographystyle{plain}
\bibliography{papers}

\end{document}